\documentclass[letterpaper, preprint, paper,11pt]{AAS}	% for preprint proceedings

\usepackage{bm}
\usepackage{amsmath}
\usepackage{subfigure}
\usepackage[colorlinks=true, pdfstartview=FitV, linkcolor=black, citecolor= black, urlcolor= black]{hyperref}
\usepackage{overcite}
\usepackage{footnpag}			      	% make footnote symbols restart on each page

\usepackage{amsmath,amsthm,amssymb,amsfonts, fancyhdr, color, comment, graphicx, environ, bbm}
\usepackage{longtable,tabularx}

\usepackage{algorithm}
\usepackage{algorithmicx}
\usepackage{algpseudocode}
\algnewcommand{\Initialize}[1]{%
  \State \textbf{Initialize:}
  \State \hspace*{\algorithmicindent}\parbox[t]{0.8\linewidth}{\raggedright #1}
}

\usepackage{graphicx}

\newcommand{\R}{\mathbbm{R}}
\newcommand{\p}{\operatorname{\mathbbm{P}}}
\newcommand{\E}{\operatorname{\mathbbm{E}}}

\newcommand{\sN}{\mathsf{N}}

\newcommand{\Var}{\operatorname{\mathsf{Var}}}

\newcommand{\bitm}{\begin{itemize}[leftmargin=*]}
\newcommand{\eitm}{\end{itemize}}
\newcommand{\benm}{\begin{enumerate}[leftmargin=*,label=(\alph*)]}
\newcommand{\eenm}{\end{enumerate}}

\PaperNumber{23-391}

\begin{document}

\title{Stochastic Hazard Detection For Landing\\ Under Topographic Uncertainty}

\author{Kento Tomita\thanks{PhD Student, School of Aerospace Engineering, Georgia Institute of Technology, Atlanta, GA 30313, USA.},  
\ Koki Ho\thanks{Associate Professor, School of Aerospace Engineering, Georgia Institute of Technology, Atlanta, GA 30313, USA.}
}
\maketitle{}

\begin{abstract}
Autonomous hazard detection and avoidance is a key technology for future landing missions in unknown surface conditions. Current state-of-the-art stochastic algorithms assume simple Gaussian measurement noise on dense, high-fidelity digital elevation maps, limiting the algorithm's applicability. This paper introduces a new stochastic hazard detection algorithm capable of more general topographic uncertainty by leveraging the Gaussian random field regression. The proposed approach enables the safety assessment with imperfect and sparse sensor measurements, which allows hazard detection operations under more diverse conditions. We demonstrate the performance of the proposed approach on the existing Mars digital terrain models. 
\end{abstract}

% TODO
% - detailed related work
% - comparison with probabilistic ALHAT
% - update figures (pipeline)
% - finer resolution images

% \section{memo}
%The treatment of safety uncertainty by the community lacks a formal framework.

%\begin{itemize}
%    \item However, it can be noted that the treatment of shape uncertainty by the small body science and engineering community lacks a formal framework. Previous works on shape reconstruction and dynamical environment characterization of small bodies do acknowledge the presence of uncertainty in the reconstructed shapes and inertia (Scheeres et al. 2006a; Nolan et al. 2013; Scheeres et al. 2016; Busch et al. 2006; Torppa et al. 2003), but do not go as far as establishing the relationship between the uncertainty in the shape and that in the inertia properties and gravity field in a systematic and quantitative way.
%\end{itemize}

\section{Introduction}
Autonomous hazard detection and avoidance (HD\&A) is a key technology for future landing missions in unknown surface conditions. Current state-of-the-art autonomous HD\&A algorithms require dense, high-fidelity digital elevation maps (DEMs), and often assume a simple Gaussian error on the range measurements~\cite{ivanov2013probabilistic}. We propose a new hazard detection (HD) algorithm capable of more general topographic uncertainty by leveraging the Gaussian random field (GRF) regression to reduce reliance on expensive, high-fidelity, dense terrain maps. 
%Current state-of-the-practice methods rely on the real-time Light Detection and Ranging (LiDAR) measurements\cite{Li2016} or \textit{a priori} safety maps generated by the satellite-imagery-based digital terrain models (DTMs).
%As the reliable topographic information, hazard detection algorithms often digital elevation maps (DEMs) obtained from point cloud data (PCD) of LiDAR measurements have been often used\cite{restrepo2020next, ivanov2013probabilistic, tomita2022jsr}. 

\begin{figure}[t]
    \centering
    \includegraphics[width=0.7\linewidth]{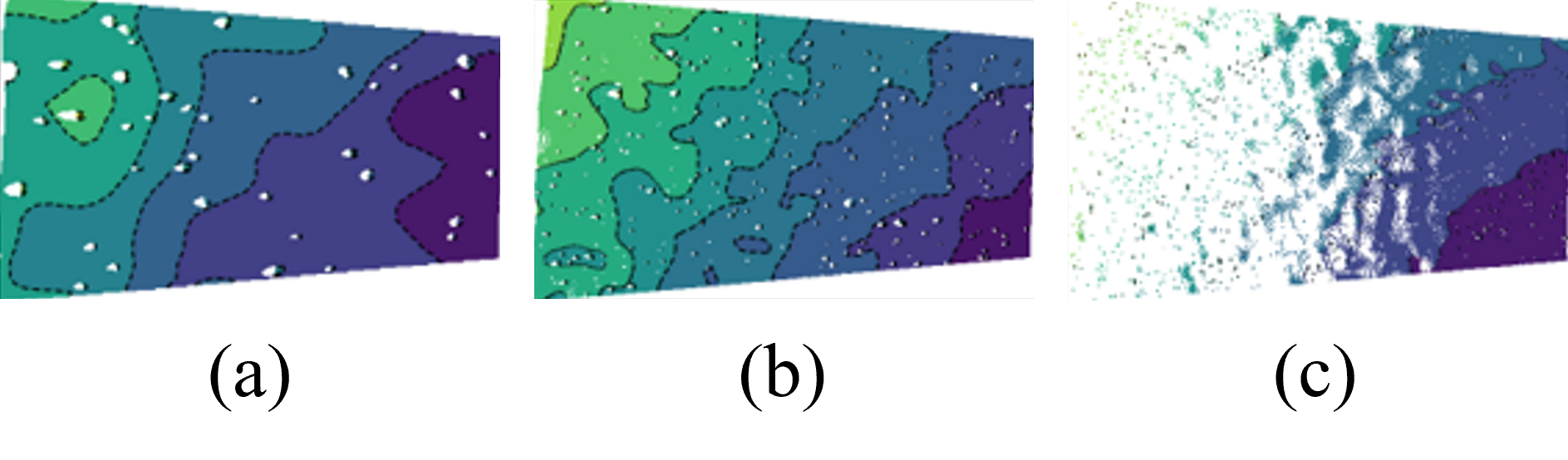}
        \caption{\textbf{Simulated DEMs over the slant ranges of (a) 100m, (d) 250m, and (c) 500m.} The LiDAR detector size is 1024x1024 and slant angle is set 30 deg. There are holes due to the rock occulusions with the large slant angle in (a) and (b) and the extended ground-sample-distances due to the large slant range and angle in (c).}
        \label{fig:lidar}
\end{figure}

Previous works~\cite{ivanov2013probabilistic, tomita2022jsr} on the uncertainty-aware HD algorithms mainly studied the effect of the range error of LiDAR sensors. 
Ivanov et al.~\cite{ivanov2013probabilistic} developed a probabilistic HD algorithm assuming the identically and independently distributed Gaussian error on the LiDAR's range measurement, given the lander geometry and navigation errors. Tomita et al.~\cite{tomita2022jsr} applied Bayesian deep learning techniques to solve the same problem with increased performance. 
However, the previous works~\cite{ivanov2013probabilistic, tomita2022jsr} assume noisy but dense DEMs, and do not incorporate the topographic uncertainty caused by the sparsity of the LiDAR measurements. %The quality of the reconstructed surface information varies depending on the surface topography, or the distance and viewing angle with respect to the surface. 
LiDAR sensors measure the distance to the terrain surface, and output point cloud data (PCD), which is the collection of the estimated coordinates of the measured points on the surface. The maximum ground sample distance (GSD) of PCD increases at higher altitudes or by non-orthogonal observations, which causes the blank spots on the obtained DEMs~\cite{restrepo2020next}. Figure \ref{fig:lidar} shows the simulated DEMs have missing data due to the large slant range or angle. To handle PCD with larger GSDs than the required DEM resolution, we leverage the Gaussian random field regression to estimate the dense DEM with appropriate uncertainty, and derive the probability of safety for each surface point.

The contributions of this paper are as follows: first, we develop a new stochastic hazard detection algorithm that leverages the Gaussian random field (GRF) to accurately predict safety probability from noisy and sparse LiDAR measurements; second, we derive the analytic form of the approximated probability of safety under the GRF representation of the terrain; and third, we demonstrate and analyze the performance of the proposed approach on real Mars digital terrain models.

%\section{Related Work}
%\input{text/related_work}

\section{Proposed Approach}\label{approach}
%%%%%%%%%%%%%%%%%%%%%%%%%%%%%%%%%%%%%%%%%%%%%%%%%%%%%%%%%%%%%

\begin{figure}[t]
    \centering
    \includegraphics[width=0.8\linewidth]{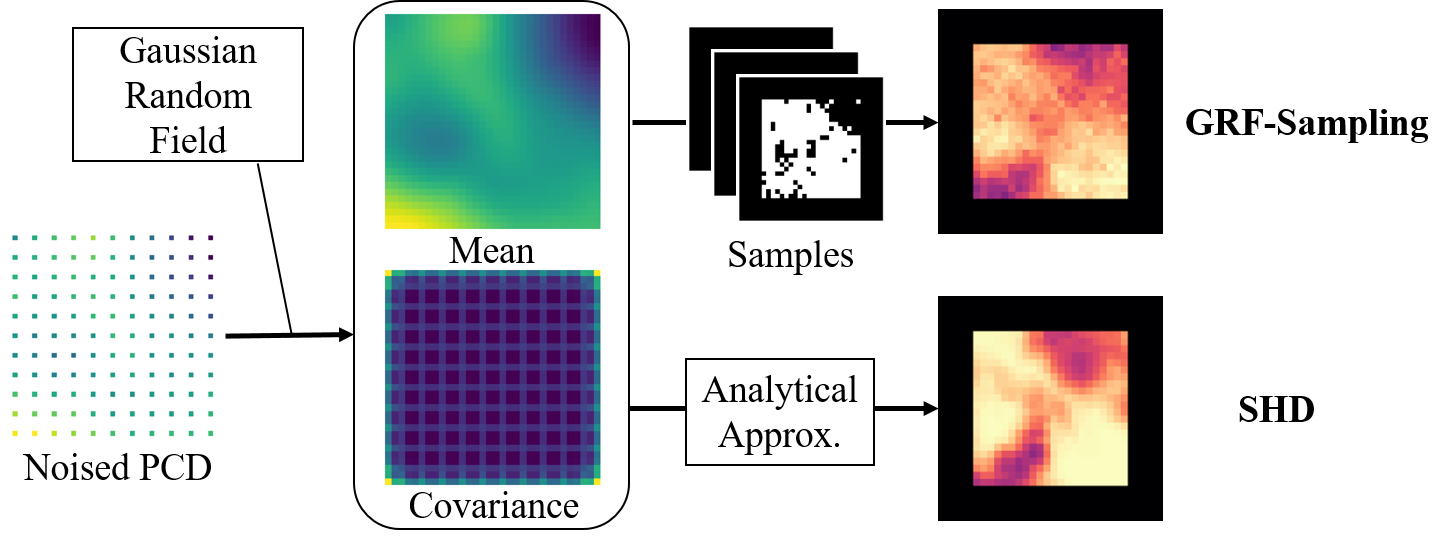}
    \caption{\textbf{Pipeline of the proposed approach.}}
    \label{fig:pipeline}
\end{figure}

\subsection{Terrain Approximation by Gaussian Random Fields}
We approximate the topography of the target terrain by a Gaussian random field (GRF), which is a joint Gaussian distribution about the surface elevations, also referred to as a two-dimensional Gaussian process~\cite{seeger2004gaussian}. Specifically, let $\gamma_i = [x_i, y_i]^T\in \mathcal{S}$ be the horizontal position of an arbitrary surface point where $\mathcal{S}\subseteq \R^2$ is the horizontal projection of all the surface points of our interest. Then, by the GRF approximation, we assume that the elevations $\{z_i = f(\gamma_i) |\gamma_i\in\mathcal{S}\}$ are jointly Gaussian such that 
\begin{equation}
    \E[f(\gamma_i)] = \mu_i, \quad \E[(f(\gamma_i)-\mu_i)(f(\gamma_j)-\mu_j)]=k(\gamma_i, \gamma_j), \quad \forall i, j = 1, 2, ..., |\mathcal{S}|
\end{equation}
for the appropriately chosen mean $\mu_i$ and the kernel function, $k(\gamma_i, \gamma_j)$. We use the absolute exponential kernel function to approximate natural terrains by a Brownian motion~\cite{malamud2001wavelet, turcotte1987fractal}
\begin{equation}
    k(\gamma_i, \gamma_j) = u \exp\left(-\frac{\|\gamma_i-\gamma_j\|}{\ell}\right)
\end{equation}\label{eq:se-kernel}
where $u>0$ and $\ell>0$ are the hyperparameters of the GRF. 

Given the terrain measurements, whether noisy or sparse, we can condition the GRF on the measurements. Suppose we have $n$ noisy observed elevations of the terrain, $\mathbf{z}=[z_1, z_2, ..., z_n]^T$, for the horizontal locations of $\Gamma=[\gamma_1, \gamma_2, ..., \gamma_n]^T$. We would like to obtain $n_*$ elevations of $\mathbf{z_*}=[z_{1_*}, z_{2_*}, ..., z_{n_*}]^T$ for $\Gamma_*=[\gamma_{1_*}, \gamma_{2_*}, ..., \gamma_{n_*}]^T$. Let $\sigma^2$ be the variance of the Gaussian observation noise, then we have the conditional distribution~\cite{seeger2004gaussian}
\begin{equation}\label{eq:gp-regression}
\begin{split}
    \mathbf{z}_*|\Gamma, \mathbf{z}, \Gamma_*&\sim\sN(\bar{\mathbf{z}}_*, \text{cov}(\mathbf{z}_*)),\quad \text{where}\\
    \bar{\mathbf{z}}_*&:=\E[\mathbf{z}_*|\Gamma, \mathbf{z}, \Gamma_*]=K(\Gamma_*, \Gamma)[K(\Gamma,\Gamma)+\sigma^2 I]^{-1}\mathbf{z}\\
    \text{cov}(\mathbf{z_*}) &= K(\Gamma_*, \Gamma_*)- K(\Gamma_*, \Gamma)[K(\Gamma, \Gamma)+\sigma^2 I]^{-1}K(\Gamma, \Gamma_*).\\
\end{split}
\end{equation}
 Here $K(\mathbf{\gamma_*}, \mathbf{\gamma})\in\R^{n_*\times n}$ is the covariance matrix whose $i,j$ entry corresponds to $k(\gamma_i, \gamma_{j_*})$, and similar for $K(\mathbf{\gamma}, \mathbf{\gamma}), K(\mathbf{\gamma}, \mathbf{\gamma_*})$, and $K(\mathbf{\gamma_*}, \mathbf{\gamma_*})$. To optimize the hyperparameters of $k$, we maximize the log marginal likelihood~\cite{seeger2004gaussian}
 \begin{equation}\label{eq:lml}
     \log p(\mathbf{z}|\Gamma) = -\frac{1}{2}\mathbf{z}^T(K+\sigma^2 I)^{-1}\mathbf{z}-\frac{1}{2}\log|K+\sigma^2 I|-\frac{n}{2}\log 2\pi
 \end{equation}
 where $K=K(\Gamma, \Gamma)$. For more details about the Gaussian random field regression, please refer to Reference~\citenum{seeger2004gaussian}.

%%%%%%%%%%%%%%%%%%%%%%%%%%%%%%%%%%%%%%%%%%%%%%%%%%%%%%%%%%%%%
\subsection{Probability of Landing Safety}

\begin{figure}[t]
    \centering
    \includegraphics[width=0.7\linewidth]{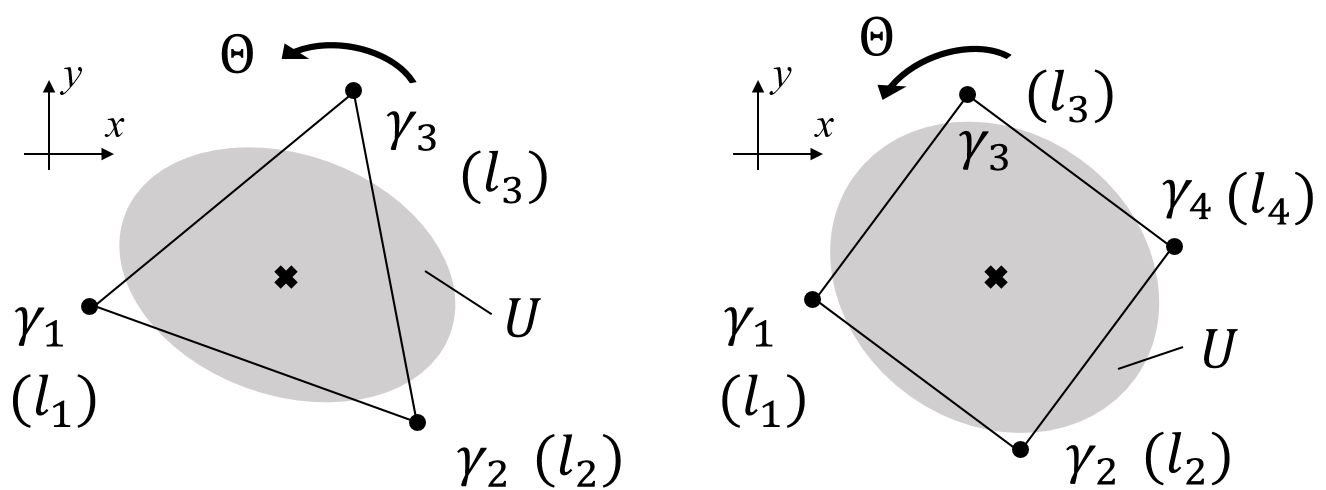}
    \caption{\textbf{Schematics of the triangular and square lander geometries.} The bold cross mark at the center is the target whose landing safety is evaluated. We evaluate all the possible landing pad placements over different orientation angles $\theta$. The gray area denoted by $U$ is the lander footprint where roughness safety is evaluated.}
    \label{fig:lander-geom}
\end{figure}

Given the GRF representation of the dense elevation map, we compute the slope and roughness at touchdown, which defines the landing safety. Suppose we have a target site whose landing safety is to be evaluated, then the slope and roughness depend on the lander's orientation angle, $\theta_o\in\Theta$, as shown in Figure \ref{fig:lander-geom}. We classify the target as safe if the slope and roughness are under the given thresholds for all the orientation angles. Therefore, the probability of safety for the target is the joint probability of $\{\text{Safe}|\theta_o\}$ over $\theta_o\in\Theta$, which is the safety given the orientation angle $\theta$. Under the GRF representation of the terrain, safety for each $\theta_o$ is not independent, and the precise computation of the joint probability is not straightforward. Instead, we approximate the joint probability by the worst-case safety probability with the raising factor of $k_1>0$, as shown in Eq. (\ref{eq:psafe}). %Suppose the lander has three landing legs, and there is a target site whose landing safety is our interest. Centering the lander at the target, the locations of the three landing pads depend on the orientation angle, $\theta$, as shown in Figure \ref{fig:lander-geom}. Given the orientation angle, as detailed later in Eq. (\ref{eq:slope-rghns}), the slope and roughness are defined as follows: the slope $s$ is the angle between the local vertical axis and the normal vector of the landing surface, which is formed by connecting the three landing pads; the roughness $r$ is the distance between the terrain and the landing surface and is evaluated over the lander footprint $U$. For the target to be safe to land, the slope and the maximum roughness over $U$ must be below the thresholds for all the orientation angles. By discretizing the orientation angles $\theta_{o}\in\Theta$, the probability of safe landing of the target is expressed as
\begin{equation}\label{eq:psafe}
\begin{split}
    \p(\text{Safe}) &=\p\{\text{Safe}|\theta_1, \text{Safe}|\theta_2, ..., \text{Safe}|\theta_{|\Theta|}\}\\
    &\sim \left[\min_{\theta_o\in\Theta}\p\{\text{Safe}|\theta_o\}\right]^{k_1}
\end{split}
\end{equation}

To compute $\p\{\text{Safe}|\theta_o\}$, let $l_1, l_2, l_3\in\R^3$ be the position of the three landing pads contacting with the terrain with the orientation angle $\theta_{o}$. 
\begin{equation}\label{eq: pad-locations}
    \begin{split}
        l_1=[x_1, y_1, z_1]^T, \quad l_2=[x_2, y_2, z_2]^T, \quad l_3=[x_3, y_3, z_3]^T
    \end{split}
\end{equation}
By the GRF approximation, $z_1, z_2,$ and $z_3$ are the random variables satisfying
\begin{equation}\label{eq:grf-zi}
    \begin{split}
        \E [z_i] =\mu_i, \quad \E [(z_i-\mu_i)(z_j-\mu_j)] = \rho_{ij}, \quad i, j=1, 2, 3
    \end{split}
\end{equation}
where $\mu_i$ and $\rho_{ij}$ are known.
We can compute the slope and roughness by finding the landing surface, which is the plane spanned by the landing pads $l_1, l_2$, and $l_3$. The normal vector, $n$, of the landing surface is obtained by taking the cross product:
\begin{equation}\label{eq:normal}
    \begin{split}
        n &= (l_2 - l_1) \times (l_3 - l_1)\\
        &= \begin{bmatrix}
            y_{12}(z_3-z_1)-y_{13}(z_2-z_1)\\
            x_{13}(z_2-z_1)-x_{12}(z_3-z_1)\\
            x_{12}y_{13} - x_{13}y_{12}
            \end{bmatrix}
            =: \begin{bmatrix}
            a\\b\\c
            \end{bmatrix}
    \end{split}
\end{equation}
where $x_{ij}=x_j - x_i$ and $y_{ij}=y_j - y_i$.
The slope $s$ is defined as the angle between the normal vector $n$ and the vertical axis, and the roughness $r(\gamma)$ is the distance between the terrain and the landing surface at the horizontal location of $\gamma\in U$, where $U$ represents the set of horizontal coordinates within the lander footprint. The slope and roughness are then computed as
\begin{equation}\label{eq:slope-rghns}
\begin{split}
    s &= \arccos{\left(\frac{c}{\sqrt{a^2 + b^2 + c^2}}\right)}\\
    \quad r(\gamma)&=\frac{|ax + by + cz + d|}{\sqrt{a^2 + b^2 + c^2}},\quad \gamma=(x, y)\in U ,\quad z=f(\gamma)
\end{split}
\end{equation}
where $z=f(\gamma)$ is the terrain elevation at $\gamma$, and $d$ satisfies $ax_i+by_i+cz_i+d=0$ for all $\ell_i=[x_i, y_i, z_i]^T$, $i=1, 2,$ and $3$. Given the slope and roughness thresholds $\bar{s}$ and $\bar{r}$, the conditional probability of safety is expressed as follows.
\begin{equation}\label{eq:theta-psafety}
    \p\{\text{Safe}|\theta_o\} = \p\{s<\bar s, \text{ and } r(\gamma)<\bar r\text{ for all }\gamma\in U|\theta_{o}\}
\end{equation}
Note that the computation of the conditional joint probability of Eq. (\ref{eq:theta-psafety}) under Eqs. (\ref{eq:grf-zi})(\ref{eq:normal})(\ref{eq:slope-rghns}) is not straightforward because $z_1, z_2, z_3$, and $z$ are all correlated random variables. To ease the computation, we approximate the probability of safe landing by decomposing it into slope safety and roughness safety, as in Eq. (\ref{eq: p-decompose}).
\begin{equation}\label{eq: p-decompose}
\begin{split}
    \p\{\text{Safe}|\theta_o\} &= \p\{s<\bar s, \text{ and } r(\gamma)<\bar r\text{ for all }\gamma\in U|\theta_{o}\}\\
    &\sim \p\{s<\bar s|\theta_o\}\p\{r(\gamma)<\bar r\text{ for all }\gamma\in U|\theta_{o}\}
\end{split}
\end{equation}
In the following subsections, we derive the analytical expressions of the conditional probability of slope safety, $\p\{s<\bar s|\theta_{o}\}$, and the conditional probability of roughness safety, $\p\{r(\gamma)<\bar r\text{ for all }\gamma\in U|\theta_{o}\}$.

%%%%%%%%%%%%%%%%%%%%%%%%%%%%%%%%%%%%%%%%%%%%%%%%%%%%%%%%%%%%%
\subsection{Probability of Slope Safety}
Here we derive the analytical form of the probability of slope safety, $\p\{s<\bar s|\theta_{o}\}$. Let us introduce the multivariate normal variable $Z_s$ to represent the joint distribution of the elevations of the three landing pads contacting the terrain with the orientation angle $\theta_{o}$.
\begin{equation*}
    Z_s := [z_1, z_2, z_3]^T \sim \sN(\mu_s,\Sigma_s),
\end{equation*}
where $\mu_s\in\R^3$ and $\Sigma_s\in\R^{3\times3}$ are known by the GRF approximation. Then, we can rewrite the conditional probability of slope safety as follows.
\begin{equation}\label{eq:psafe_slope}
    \begin{split}
        \p\{s<\bar s|\theta_{o}\} &= 
        \p\left\{a^2+b^2 < c^2\left(\frac{1}{\cos^2(\bar s)}-1\right)\right\}\\
        &=\p\left\{Z_s^T A(\theta_{o}) Z_s < \tau_{\bar{s}}(\theta_{o})\right\},\\\text{where}&\\
        A(\theta_{o}) &= 
         \begin{bmatrix}
            x_{23}^2 + y_{23}^2 & -x_{13}x_{23}-y_{13}y_{23} & x_{12}x_{23}+y_{12}y_{23}\\
            -x_{13}x_{23}-y_{13}y_{23} & x_{13}^2+y_{13}^2 & -x_{12}x_{13}-y_{12}y_{13}\\
            x_{12}x_{23}+y_{12}y_{23} & -x_{12}x_{13}-y_{12}y_{13} & x_{12}^2+y_{12}^2\\
        \end{bmatrix},\\
        \tau_{\bar{s}}(\theta_{o})&=c^2\left(\frac{1}{\cos^2(\bar s)}-1\right)
    \end{split}
\end{equation}
Note that both $A(\theta_{o})$ and $\tau_{\bar{s}}(\theta_{o})$ are constant and depends on the lander's orientation angle $\theta_{o}$ at touchdown. However, if we ignore the roughness safety and are only interested in the slope safety, we can make $A$ invariant over $\theta_o$ by fixing the $xy$-coordinate with respect to the horizontal locations of the three landing pads, $\gamma_1, \gamma_2$, and $\gamma_3$.

The probability of slope safety, derived as Eq. (\ref{eq:psafe_slope}), is represented as the tail distribution of the quadratic form of the multivariate normal distribution. The quadratic form of the multivariate normal distribution is known as the generalized chi-squared distribution, which is a linear combination of independent non-central chi-square variables~\cite{rencher2008linear}. Here we approximate the quadratic form of the multivariate normal distribution by the Gaussian distribution with the mean and the variance obtained as follows~\cite{rencher2008linear}.
%https://en.wikipedia.org/wiki/Quadratic_form_(statistics)
\begin{equation}
    \begin{split}
        m_s(\theta_{o}) &:= \E \left[Z_s^T A Z_s\right] =\text{tr}(A\Sigma_s) + \mu_s^TA\mu_s\\
        \sigma_s^2(\theta_{o})&:=\Var \left[Z_s^T A Z_s\right] = 2\text{tr}(A\Sigma_s A\Sigma_s) + 4\mu_s^TA\Sigma_s A\mu_s
    \end{split}
\end{equation}
Then, by using the cumulative distribution function for the standard normal distribution, $\Phi(\cdot)$, we can approximate the conditional probability of slope safety by
\begin{equation}\label{eq:psafe_slope_cdf}
\begin{split}
\p\{s<\bar s|\theta_{o}\} &\sim\Phi\left(\frac{\tau_{\bar{s}}(\theta_{o}) - m_s(\theta_{o})}{\sqrt{2}\sigma_s(\theta_{o})}\right). % = \frac{1}{2}\left(1 + \text{erf}\left(\frac{\tau - m_s}{\sigma_s\sqrt{2}}\right)\right)
\end{split}
\end{equation}

%%%%%%%%%%%%%%%%%%%%%%%%%%%%%%%%%%%%%%%%%%%%%%%%%%%%%%%%%%
\subsection{Probability of Roughness Safety}
The precise computation of the conditional probability of roughness safety requires evaluating integrals for all the points beneath the lander. To ease the computation, we introduce the rasing factor $k_2>0$ to approximate the conditional probability by the worst-case roughness instance within the landing footprint as in Eq. (\ref{eq:rghns-psafe}).
\begin{equation}\label{eq:rghns-psafe}
\begin{split}
    \p\{r(\gamma)<\bar r,\text{ for all }\gamma\in U|\theta_{o}\} &= \int_{-\infty}^{\bar{r}} \cdot\cdot\cdot\int_{-\infty}^{\bar{r}}\p(r_1,..., r_{|U|}|\theta_{o})dr_1\cdot\cdot\cdot dr_{|U|}\\
    &\sim \left[\min_{\gamma_p\in U}\int_{-\infty}^{\bar{r}}\p(r(\gamma_p)|\gamma_p, \theta_{o})dr_p\right]^{k_2}\\
    &=\left[\min_{\gamma_p\in U}\p\{r(\gamma_p)<\bar{r}|\gamma_p,\theta_{o}\}\right]^{k_2}, \quad
    \text{where }
    k_2>0
\end{split}
\end{equation}

Since $0<\Phi(\cdot)<1$, larger $k_2$ corresponds to a conservative approximation. 
The proposed approximation with $k_2=1$ represents the case when the worst-case roughness being under the threshold guarantees the roughness safety for the other points within $U$. When $k_2=|U|$, it represents the case where all the points in $U$ are independent and have an equally large roughness.

Similarly to the slope case, let us introduce the multivariate normal variable $Z_r=[z_1, z_2, z_3, z_p]^T$ to represent the joint distribution of the four elevations; $z_1, z_2,$ and $z_3$ are the elevations of the three landing pads contacting the terrain surface with orientation angle $\theta_o$, and $z_p=f(\gamma_p)$ is the elevation of an arbitral point within the lander footprint such that $\gamma_p=[x_p, y_p]^T\in U$. Then $Z_r$ follows the multivariate normal distribution, %Suppose the worst-case roughness instance is denoted by the superscript of $^*$, i.e., $\gamma^* =[x^*, y^*]^T$ and $z^* = f(\gamma^*)$, then $Z_r$ is defined as 
\begin{equation*}
    Z_r := [z_1, z_2, z_3, z_p]^T \sim \sN(\mu_r,\Sigma_r),
\end{equation*}
where $\mu_r\in\R^4$ and $\Sigma_r\in\R^{4\times4}$ are known by the GRF approximation. With $Z_r$, we can rewrite the conditional probability of roughness safety for an arbitral point $\gamma_p\in U$, as follows.
\begin{equation}\label{eq:psafe_rghns}
    \begin{split}
        \p\{r(\gamma_p)<\bar r|\gamma_p,\theta_{o}\} &= 
        \p\left\{(ax_p + by_p + cz_p +d)^2 < \bar{r}^2(a^2+b^2+c^2)\right\}\\
        &=\p\left\{(ax_p + by_p + cz_p - ax_1 - by_1 - cz_1)^2 < \bar{r}^2(a^2+b^2+c^2)\right\}\\
        &=\p\left\{(ax_{1p} + by_{1p} + c(z_p-z_1))^2 - \bar{r}^2(a^2+b^2) < \bar{r}^2c^2)\right\}\\
        &=\p\left\{Z_r^T B(\gamma_p,\theta_{o}) Z_r < \tau_{\bar{r}}(\theta_{o})\right\},\\\text{where}&\\
        \tau_{\bar{r}}(\theta_{o})&=\bar{r}^2c^2\\
        B(\gamma_p,\theta_{o})&=[\beta_{ij}] \\
        \beta_{11} &= (x_{1p}y_{23}-x_{23}y_{1p})^2-\bar{r}^2(x_{23}^2+y_{23}^2)+c^2-2c(x_{1p}y_{23}-x_{23}y_{1p})\\
        \beta_{22} &= (x_{1p}y_{13}-x_{13}y_{1p})^2-\bar{r}^2(x_{13}^2+y_{13}^2)\\
        \beta_{33} &= (x_{1p}y_{12}-x_{12}y_{1p})^2-\bar{r}^2(x_{12}^2+y_{12}^2)\\
        \beta_{44} &= c^2\\
        \beta_{12}=\beta_{21}&= -(x_{1p}y_{13}-x_{13}y_{1p})(x_{1p}y_{23}-x_{23}y_{1p}-c)+\bar{r}^2(x_{13}x_{23}+y_{13}y_{23})\\
        \beta_{13}=\beta_{31}&= (x_{1p}y_{12}-x_{12}y_{1p})(x_{1p}y_{23}-x_{23}y_{1p}-c)-\bar{r}^2(x_{12}x_{23}+y_{12}y_{23})\\
        \beta_{14}=\beta_{41} &= c(x_{1p}y_{23}-x_{23}y_{1p} - c)\\
        \beta_{23}=\beta_{32}&=-(x_{1p}^2 - \bar{r}^2)y_{12}y_{13} -( y_{1p}^2 - \bar{r}^2)x_{12}x_{13} + x_{1p}y_{1p}(x_{12}y_{13}+x_{13}y_{12})\\
        \beta_{24}=\beta_{42}&=-c(x_{1p}y_{13}-x_{13}y_{1p})\\
        \beta_{34}=\beta_{43}&= c(x_{1p}y_{12}-x_{12}y_{1p}).\\
    \end{split}
\end{equation}
We used the relation $ax_1+by_1+cz_1+d=0$ to erase $d$.
Note that both $B(\gamma_p,\theta_{o})$ and $\tau_{\bar{r}}(\theta_{o})$ are constant. 

Analogously to the slope safety, we approximate the derived conditional probability by its mean and variance. Finally, we obtain the following approximation about the conditional probability of roughness safety. 
\begin{equation}\label{eq:psafe_rghns_cdf}
\begin{split}
    \p\{r(\gamma_p)<\bar r|\gamma_p,\theta_{o}\} &\sim\Phi\left(\frac{\tau_{\bar{r}}(\theta_{o}) - m_r(\theta_{o})}{\sqrt{2}\sigma_r(\theta_{o})}\right),\quad\text{where}\\
    m_r(\theta_{o}) &:= \E \left[Z_r^T B Z_r\right] =\text{tr}(B\Sigma_r) + \mu_r^TB\mu_r\\
        \sigma_r^2(\theta_{o})&:=\Var \left[Z_r^T B Z_r\right] = 2\text{tr}(B\Sigma_r B\Sigma_r) + 4\mu_r^TB\Sigma_r B\mu_r
\end{split}
\end{equation}

%%%%%%%%%%%%%%%%%%%%%%%%%%%%%%%%%%%%%%%%%%%%%%%%%%%%%%%%%%%%%
\subsection{Stochastic Hazard Detection Algorithm}
Algorithm \ref{alg:shd} shows the resulting stochastic hazard detection algorithm that takes the GRF representation of the approximated terrain, and returns the probability of slope safety and the probability of roughness safety for each point on the terrain. For efficient implementation, the matrices $A(\theta_o)$ and $B(\gamma_p, \theta_o)$ in Eqs. (\ref{eq:psafe_slope})(\ref{eq:psafe_rghns}) should be precomputed. Note that $A$ and $B$ can be reused for different targets by taking the target-centered coordinates, so the number of $A$ and $B$ matrices to be precomputed are $|\Theta|$ and $|U||\Theta|$, respectively. %Note that $A$ is invariant for different $\theta_o$ if the three landing pad horizontal locations $\gamma_1, \gamma_2, \gamma_3$ are point symmetric, i.e., a triangular lander with the circular lander footprint. In such cases, the number of $A$ and $B$ matrices to be precomputed are $1$ and $|U|$, respectively. 

If we have a fourth landing pad, we check if the fourth landing pad is above the landing surface; given the fourth landing pad location, $\l_4=[x_4, y_4, z_4]$, we skip $\theta_o$ if $z_4 < (-ax_4-by_4-d)/c$. Although $a, b, d$, and $z_4$ are correlated random variables, we can use the expected value for the approximated feasibility check. To be conservative, we can replace $z_4$ as $z_4\leftarrow \E[z_4] - 3\sqrt{\Var[z_4]}$.
 
\begin{algorithm}
    \caption{Stochastic Hazard Detection}
    \label{alg:shd}
    \begin{algorithmic}
      \Require GRF representation of terrain
      \Ensure Probabilistic safety map
      \State{Precompute $A$ matrices of Eq. (\ref{eq:psafe_slope}) for all $\theta_o\in\Theta$}
      \State{Precompute $B$ matrices of Eq. (\ref{eq:psafe_rghns}) for all $\gamma_p\in U$ and $\theta_o\in\Theta$}
      \For{Targets in DEM}
      \For{Orientation angles $\theta_o\in\Theta$}
      \State{Evaluate $\p\{s<\bar s|\theta_{o}\}$ by Eq. (\ref{eq:psafe_slope_cdf})}
      \State{Update $\min_{\theta_o\in\Theta}\left[\p\{s<\bar s|\theta_{o}\}\right]$ for the target}
      \For{Terrain under lander footprint $\gamma_p\in U$}
      \State{Evaluate $\p\{r(\gamma_p)<\bar r|\gamma_p,\theta_{o}\}$ by Eq. (\ref{eq:psafe_rghns_cdf})}
      \State{Update $\min_{\gamma_p\in U, \theta_o\in\Theta}\left[\p\{r(\gamma_p)<\bar{r}|\gamma_p, \theta_o\}\right]$ for the target}
      \EndFor
      \EndFor
      \State{Store $\p^*\{s<\bar s\} := \min_{\theta_o\in\Theta}\left[\p\{s<\bar s|\theta_{o}\}\right]$ for the target}
      \State{Store $\p^*\{r<\bar r\} :=\min_{\gamma_p\in U, \theta_o\in\Theta}\left[\p\{r(\gamma_p)<\bar{r}|\gamma_p, \theta_o\}\right]$ for the target}
      \EndFor
    \State{Adjust the probabilities by the rasing factors $k_1$ and $k_2$ with Eqs. (\ref{eq:psafe})(\ref{eq:rghns-psafe}) }
    \end{algorithmic}
\end{algorithm}

\section{Experiments and Results}
\subsection{Experiment Configurations}
We evaluated the proposed algorithm on the HiRISE digital terrain model (DTM) of the candidate ExoMars landing site in Hypanis Valles~\cite{McEwen2007}. We cropped the DTM into 100 DEMs with the size of 32x32 at the maximum resolution of 1 meter per pixel (mpp). We evaluated these DEMs to obtain the true slope and roughness value per pixel. Here we assumed a triangular geometry of the lander with the diameter of $10$ meters.
To simulate the sparse LiDAR measurements, we downsampled the DEMs with the GSD of 1.5, 2, 3, 4 meters with the Gaussian elevation noise $\epsilon\sim\sN(0, \sigma^2)$ where $3\sigma = 5cm$. The simulated sparse LiDAR measurements are upsampled by fitting the GRF regressor of Eq. (\ref{eq:gp-regression}). 

Given the sparse measurements and the associated GRFs, We evaluated the slope and roughness safety by three different models: the baseline model, the GRF-sampling model, and the stochastic hazard detection (SHD) model. The baseline model reconstructs the high-resolution DEMs via bilinear interpolation, and deterministically measure the slope and roughness. The GRF-sampling model numerically samples 100 high-resolution DEMs from the GRF, and obtain the probability of safety as the sample mean of the deterministic evaluations. The SHD model takes the GRF and analytically approximate the probability of safety, as described in the previous section.

\begin{table}[hb]
\centering
\caption{\textbf{Optimal raising factors of Eqs. (\ref{eq:psafe})(\ref{eq:psafe_rghns}).}}
\begin{tabular}{p{20mm}p{10mm}p{10mm}p{10mm}p{10mm}}
\hline
 \textbf{Raising} & \multicolumn{4}{c}{\textbf{GSD}}\\
 \textbf{Factor}& \textbf{$1.5 m$} &  \textbf{$2.0 m$} & \textbf{$3.0 m$} &  \textbf{$4.0 m$} \\ \hline \hline
$k_1$ & 3.00 & 3.64 & 4.91 & 5.91 \\ 
$k_2$ & 0.42 & 0.74 & 1.04 & 1.09 \\ 
$k_1 k_2$ & 1.27 & 2.71 & 5.11 & 6.45 \\ \hline \hline
\end{tabular}
\label{table:k}
\end{table}

\begin{figure}[hb]
    \centering
    \includegraphics[width=0.8\linewidth]{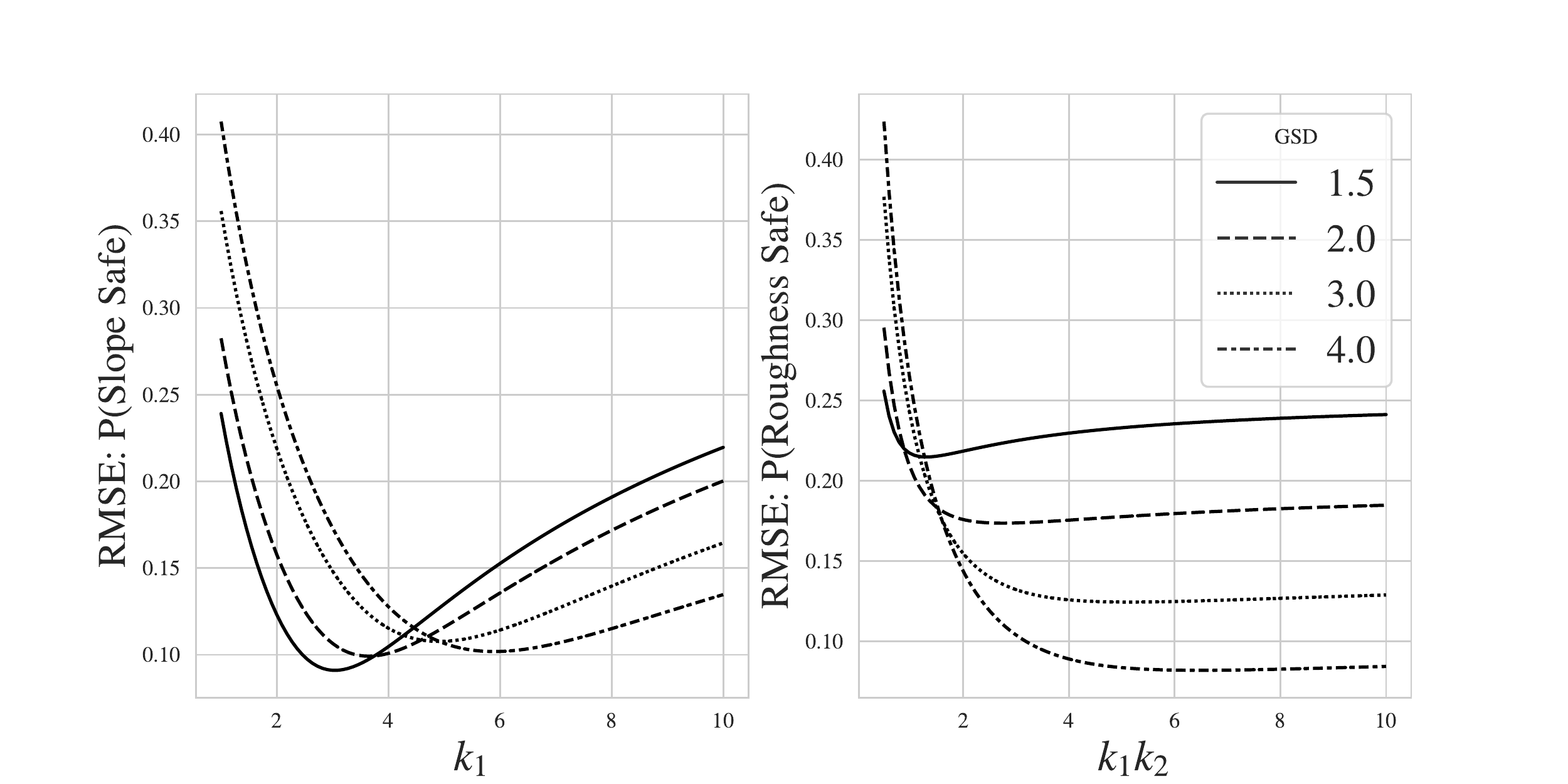}
    \caption{\textbf{Approximation errors of the proposed analytical forms: Eqs. (\ref{eq:psafe})(\ref{eq:psafe_slope_cdf})(\ref{eq:psafe_rghns_cdf}).} GSD is in meters.}
    \label{fig:rmse-k}
\end{figure}
\subsection{Accuracy of Derived Analytical Probabilities}
% RMSE for analytical approximation performance
To capture the approximation errors of the proposed analytical forms of Eqs. (\ref{eq:psafe})(\ref{eq:psafe_slope_cdf})(\ref{eq:psafe_rghns_cdf}), Figure \ref{fig:rmse-k} shows the root-mean-squared error (RMSE) of the probabilities of slope safety and roughness safety, between the analytical forms and the sample means over different raising factors $k_1$ and $k_1 k_2$. Note that the probability of roughness safety has the raising factor of $k_1 k_2$ due to Eqs. (\ref{eq:psafe})(\ref{eq:psafe_rghns_cdf}). The proposed analytical forms achieve the minimum RMSE of about 0.10 for the probability of slope safety, and from about 0.05 to 0.32 for the probability of roughness safety. As GSD increases, the optimal raising factors $k_1$ and $k_2$ increase for both slope and roughness probabilities, and RMSE of roughness-safe probability decreases. As shown later, this is related to the results that the probability of safety gets smaller for larger GSD due to the increased uncertainty, and the mean probability of roughness-safety approaches close to zero.  Table \ref{table:k} reports the optimal raising factors minimizing RMSE. We used the optimal rasing factors for the following results.

\begin{figure}[h]
    \centering
    \includegraphics[width=0.85\linewidth]{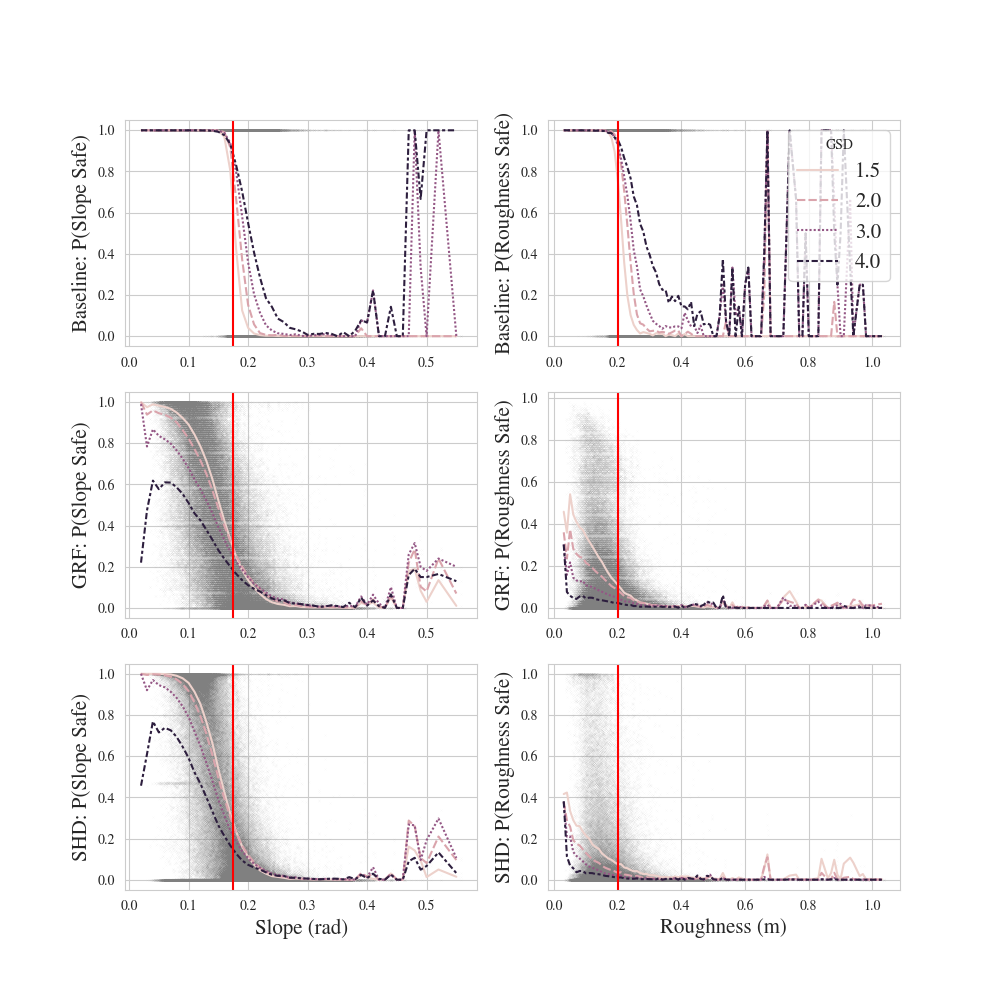}
    \caption{\textbf{Distribution of the estimated safety probability by the baseline, the GRF-sampling model, and the SHD model, from top to bottom.} Vertical red lines denote the safety thresholds. The line plots show the mean estimated safety probabilities given. The baseline is a deterministic model, and the estimated safeties are located at either 1 or 0.}
    \label{fig:dist}
\end{figure}

% Model comparison: distribution
\subsection{Prediction Performance of Proposed Approach}
Figure \ref{fig:dist} shows the distribution of the estimated safety probability by the baseline, the GRF-sampling model, and the SHD model, from top to bottom. Line plots show the mean estimated probability for different GSD cases. The baseline model has estimated safety probabilities of either 1 or 0, as it is a deterministic algorithm. The baseline algorithm fails to detect hazards from noisy, sparse inputs, resulting in the larger mean probabilities of safety on the right of the vertical red lines, which denotes the safety thresholds. Higher GSD inputs result in more missed hazards, represented by the increased safety probability in the region over the thresholds. 

On the other hand, the GRF-sampling model and the SHD model successfully assign lower safety probabilities than the baseline to hazardous targets. We can also observe the distributions of the GRF-sampling and SHD models overlaps, showing the precision of the derived analytical approximations. Note that the raising factors $k_1$ and $k_2$ are constant over the same GSD inputs, and their optimization cannot arbitrarily change the distribution; it only compresses the SHD distributions vertically.

The GRF-sampling model and the SHD model both assign larger safety probabilities to less hazardous targets, and their safety probability decreases for larger input GSDs due to the increased uncertainty. The mean estimated probabilities of roughness safety are kept relatively low even for the safe targets, and approach close to zero with larger GSDs. This means roughness safety is more sensitive to the topographic uncertainty than slope safety.

% Qualitative comparison
\subsection{Qualitative Safety Mapping Results}
\begin{figure}[h]
    \centering
    \includegraphics[width=0.8\linewidth]{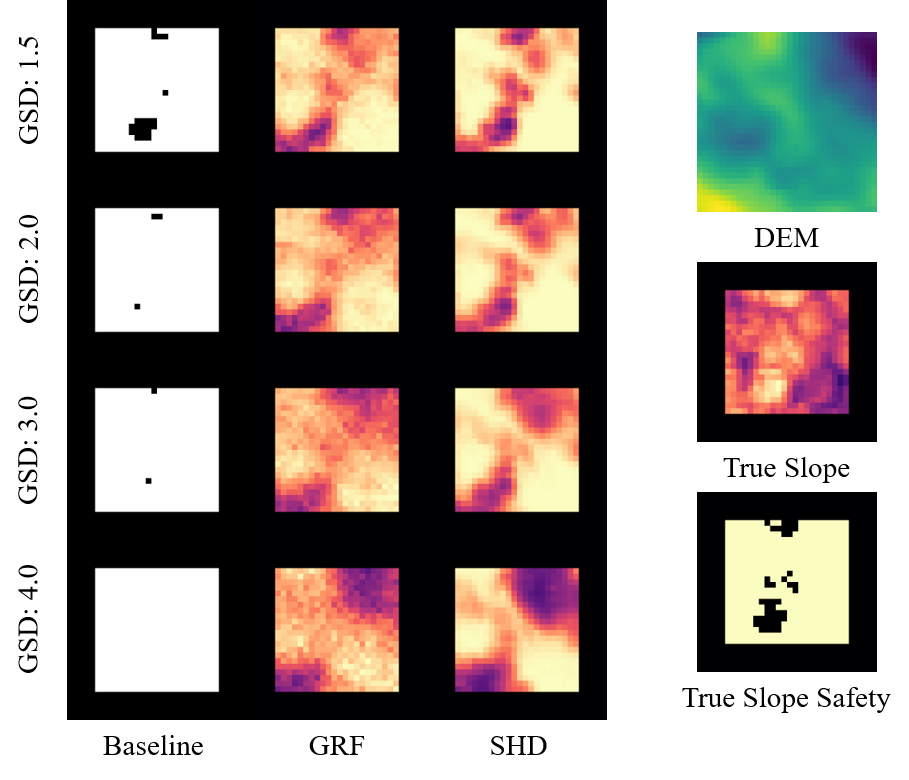}
    \caption{\textbf{Estimated slope safety maps.} The brighter pixels correspond to safe targets for the true safety map, and the estimated safety probability for the baseline, GRF-sampling, and SHD model predictions. The brighter pixels of the DEM and the true slope map correspond to their higher values.}
    \label{fig:slope-maps}
\end{figure}
\begin{figure}[h]
    \centering
    \includegraphics[width=0.8\linewidth]{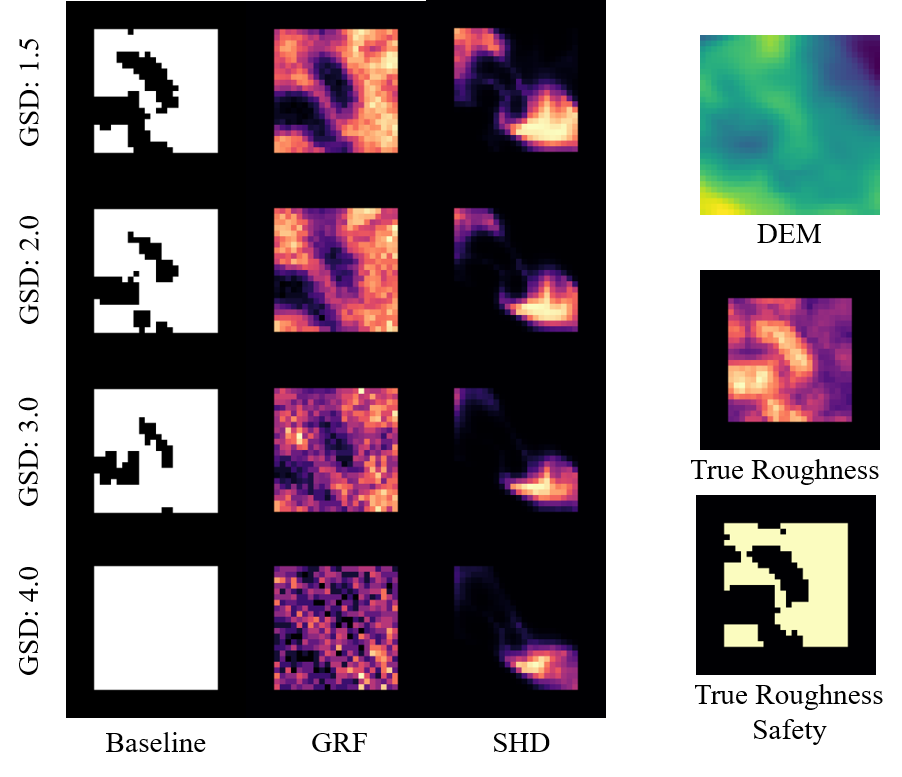}
    \caption{\textbf{Estimated roughness safety maps.} The brighter pixels correspond to safe targets for the true safety map, and the estimated safety probability for the baseline, GRF-sampling, and SHD model predictions. The brighter pixels of the DEM and the true roughness map correspond to their higher values.}
    \label{fig:rghns-maps}
\end{figure}

Figures \ref{fig:slope-maps} and \ref{fig:rghns-maps} show the estimated safety probability maps for slope and roughness, respectively. The baseline predictions miss landing hazards, denoted by black pixels, especially for higher GSD inputs. Compared to the baseline, the GRF-sampling and SHD models successfully capture the landing hazards even for higher GSD inputs, which illustrates that the GRF-sampling and SHD models allow successful hazard detection operations from noisy, sparse terrain observations.

Comparing the GRF and SHD models, the SHD model is less noisy and has more sharp boundaries of safety, especially for the larger GSD inputs. This is because the SHD model is based on the analytical expressions of estimated probability, instead of the safety samples as in the GRF-sampling model; for the larger GSD inputs, the increased uncertainty decreases the sample efficiency with respect to the precise probability estimation. However, the analytical expressions of SHD can minimize this effect and enables less noisy probability estimations.

\section{Conclusion}
We proposed a new stochastic hazard detection (HD) algorithm capable of more general topographic uncertainty by leveraging the Gaussian random field (GRF) regression. Given the noisy, sparse topographic observations, we demonstrated the GRF-based HD algorithm can detect landing hazards that are missed by the bilinear-interpolation-based algorithm. Further, we derived the analytical approximations of the safety probability and demonstrated the accuracy of the derived expressions. The numerical experiments with the existing Mars terrain model showed the analytical evaluation of the safety probability is more robust to the increased topographic uncertainty than the sampling based algorithm. We demonstrated that the proposed approach enables the safety assessment with imperfect and sparse sensor measurements, which allows hazard detection operations under more diverse conditions.

\section*{Acknowledgments}
This work is supported by the National Aeronautics and Space Administration under Grant No.80NSSC20K0064 through the NASA Early Career Faculty Program.

\bibliographystyle{AAS_publication}   % Number the references.
\bibliography{references}  % Use references.bib to resolve the labels.

\end{document}